\documentclass{article}
\usepackage[dblblindworkshop,final,nonatbib]{neurips_2026}
\workshoptitle{AgenticLS: Agentic AI for Biological Discovery}
\makeatletter
\renewcommand{\@noticestring}{Submitted to AgenticLS 2026: Agentic AI for Biological Discovery, a NeurIPS 2026 Workshop. Do not distribute.}
\makeatother
\usepackage[utf8]{inputenc}
\usepackage[T1]{fontenc}
\usepackage[hidelinks]{hyperref}
\usepackage{booktabs,array,amsmath,amssymb,microtype,graphicx,caption,enumitem}
\title{Identity or Evidence? Auditing Evidence Dependence\\ in Agentic Life-Science Benchmarks}
\author{
  Danai Brilli\\
  Zo Labs\\
  \texttt{dbrilli@zolabs.bio}
  \And
  Maria Georgila\\
  Zo Labs\\
  \texttt{mgeorgila@zolabs.bio}
}
\begin{document}
\maketitle
\begin{abstract}
Agentic life-science benchmarks give an agent evidence about a named entity and score an answer against a recorded outcome. That score tests reasoning over evidence only if the evidence, rather than the name, determines the answer. We examine this distinction by varying identity and evidence independently on a clinical-outcome benchmark and published tasks from Biomni and LAB-Bench. Blinding identity matters little when a prompt contains a decisive sequence or passage. It matters a great deal when the prompt is mostly a name and candidate list (0.742 named versus 0.082 blinded; chance 0.093). Across 96 LAB-Bench LitQA2 questions, the gap increases under one fixed passage-redaction scheme (+0.000 to +0.135; per-instance slope +0.147, $p=0.004$). On the clinical benchmark, a no-evidence recall probe matches prediction accuracy and recovers a database-specific programme count. So, we argue that benchmarks intended to measure evidence-dependent reasoning should report identity-blinded results, retrieval telemetry, and positive controls.
\end{abstract}
\begin{figure}[!h]\centering
\includegraphics[width=0.88\textwidth]{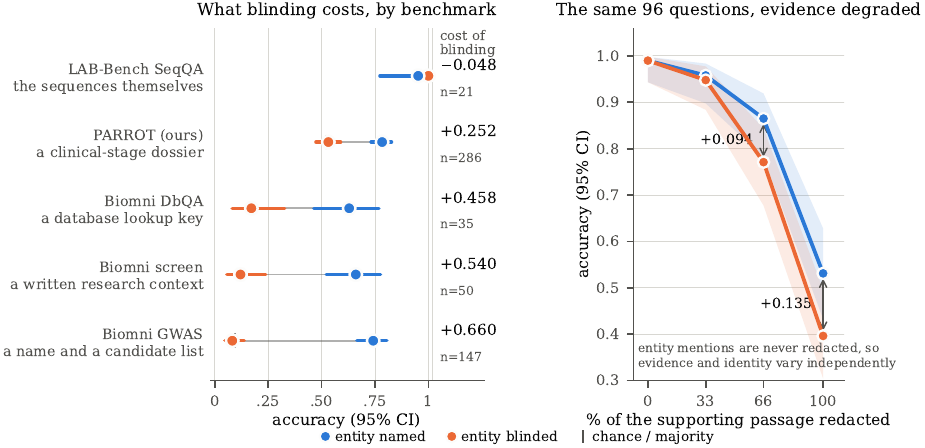}
\caption{\textbf{The cost of identity blinding varies with available evidence.} Left: conditions within a row differ only in entity mentions; numeric evidence is byte-identical. Right: on 96 LitQA2 questions, the supporting passage is redacted in four steps while entity mentions remain intact. The blinding gap grows with redaction (per-instance slope +0.147, $p=0.004$).}
\label{fig:teaser}\end{figure}
\section{Introduction}
Agentic biology benchmarks reconstruct evidence available at a scientific decision point, remove the conclusion, and score an agent against the eventual recorded outcome. This design spans molecular discovery, genomics, single-cell analysis, and laboratory tasks \cite{txbencholigo,txbenchpp,txbenchab,variantbench,scbench,spatialbench,epibench,bixbench,scienceagentbench,corebench,labbench,biomni,biodiscovery}. Its intended construct is reasoning from supplied evidence. But if a prompt retains a recognisable gene, drug, disease, or study, an agent may instead use prior knowledge or retrieve the recorded answer.

We ask a simple question: how much of the score survives when the name disappears and the evidence stays put? The answer differs across the tasks we audited. Identity is largely irrelevant when the prompt itself supplies the needed evidence, but it becomes important when it does not. 
\paragraph{Contributions.} The paper introduces paired prompt variants that separate identity, evidence, and record recall. It then uses LitQA2 to show that the cost of hiding identity rises as passage evidence is removed. Finally, it shows why deleting evidence alone is a poor test of content use: deletion changes the prompt in several ways at once.

\begin{table}[t]\centering\footnotesize
\caption{Interpretive status of the main analyses.}\label{tab:status}
\begin{tabular}{@{}p{2.0cm}p{1.3cm}p{4.1cm}p{3.5cm}@{}}\toprule
analysis & status & evidence & key limitation \\ \midrule
LitQA2: full passage & primary & 0.990 named and blinded at 0\% redaction & 96 questions; one model \\
LitQA2: redaction gradient & primary & identity-gap slope +0.147, $p=0.004$ & one seeded mask \\
Clinical dossier: blind & supporting & 0.531 blinded versus 0.783 as written & semantic context also removed \\
Clinical dossier: recall & supporting & 0.818 all-case accuracy; count $\rho=0.654$ & not a per-prediction mechanism test \\
Deletion controls & diagnostic & controls are inconclusive & no attribution is claimed \\
\bottomrule\end{tabular}\end{table}

\section{Related work and protocol}
Partial-input baselines have exposed hypothesis-only inference and passage-free comprehension \cite{artifacts,poliak,kaushik}, knowledge-conflict and contamination probes test related failures \cite{longpre,sainz}. Feng et al.\ \cite{feng} make the complementary point: changing an input does not, by itself, identify what caused a model's response. We treat this as a discriminant-validity problem \cite{jacobs}, drawing on work about benchmark scrutiny, confidence, irrelevant context, and statistical power \cite{bowman,kapoor,kadavath,xiong,shi,lostmiddle,power}.

From an instance with evidence $E$, identity $I$, and outcome, the audit creates paired variants. \textsc{blind} holds $E$ fixed while removing $I$. \textsc{delete} holds $I$ fixed while removing $E$; \textsc{placebo} replaces $E$ with another instance's evidence. \textsc{recall} is run only after blinding bites, to ask whether identity gives access to the recorded label. Positive controls verify that the instrument can transmit a decisive change. Every contrast is paired over the same instances and reported with an exact McNemar test, Wilson interval, and minimum detectable effect (MDE) for nulls \cite{power}. The LitQA2 full-passage comparison and redaction slope are primary analyses, while clinical blind and recall are supporting evidence. Full prompts, raw responses, and scripts are supplementary.

\paragraph{What each contrast estimates.} Let $A(E,I)$ denote paired accuracy under evidence $E$ and identity $I$. The identity cost is $A(E,I)-A(E,\varnothing)$: it is positive when the named prompt outperforms an otherwise identical blinded prompt. The evidence contrast is $A(E,I)-A(\varnothing,I)$, but it is not a content-use estimand by itself: removing a block also changes length, formatting, and the option to use that block. A placebo comparison is more informative only when the replacement has known signal. Thus a blinded contrast can establish dependence on the disclosed identity under this intervention. It cannot decide whether that dependence reflects memorised records, biological knowledge, or the loss of semantic context. The recall probe and functional-description diagnostic help distinguish these explanations to some extent.

\begin{table}[t]\centering\footnotesize
\caption{Core audit conditions. $E$ is the evidence block and $I$ the entity identity. All contrasts are paired, while the synthetic positive control measures the instrument, not the benchmark.}\label{tab:arms}
\begin{tabular}{@{}lp{4.2cm}p{5.8cm}@{}}\toprule
condition & what it changes & what a movement means \\ \midrule
AS-WRITTEN & nothing & the benchmark's own number, under our harness \\
BLIND & $I\rightarrow$ opaque salted tokens & does the score survive without the name? \\
DELETE & $E\rightarrow$ a note that there is none & does the score use the evidence at all? \\
PLACEBO & $E\rightarrow$ another instance's evidence & if this costs what deletion costs, not content \\
RECALL & $E$ and the question cut; record asked for & does the agent already know the label? \\
POSITIVE & a decisive line, right 95\% of the time & can the instrument detect evidence? \\
\bottomrule\end{tabular}\end{table}

\section{Experimental setup}
\paragraph{Benchmark and label.} From Minikel et al.\ \cite{minikel,nelson}, we construct 286 target--indication instances (286 genes, 183 indications; prevalence 0.521), stratified by outcome, genetic support, and prominence. Each dossier contains association counts, MeSH similarity, provenance, years, locus-to-gene scores, allele frequencies, and twelve similar associations,while outcome-derived fields are excluded by assertion. The task is whether a programme completed Phase II advances beyond it. Genetically supported pairs are over-sampled, so we make no population-level claim. The label is a snapshot-level, censored highest recorded phase, not a measure of scientific decision quality: it aggregates programmes, labels ``not advanced'' rather than failure, and has no per-programme date.

\paragraph{Field, signal, and execution checks.} Four rendered labels are materially less precise than their source quantities: ``AF'' is ALT-allele frequency rather than minor-allele frequency; ``fine-mapping share'' is a normalised Open Targets L2G score rather than a posterior, exact MeSH identity is not biological relatedness and 174 of 801 counted associations bypass the displayed strict-support threshold. These issues do not alter paired contrasts with the same evidence block, but limit interpretation of deletion and placebo conditions.

The rendered \texttt{Rare/orphan indication} field is a programme-level regulatory designation, not an intrinsic property of an indication. It reproduces most of the simple supervised baseline, whereas genetics alone has little signal:
\begin{center}\small\begin{tabular}{@{}lrr@{}}\toprule
features (gene-grouped 5-fold CV) & population & 286-pair subset \\ \midrule
majority class & 0.569 & 0.523 \\
orphan flag only & 0.600 & 0.582 \\
17 genetics features, no orphan flag & 0.579 & 0.495 \\
genetics plus orphan flag & 0.610 & 0.488 \\
\bottomrule\end{tabular}\end{center}

We use \texttt{claude-opus-5} and repeat selected conditions on a smaller model from the same family. Grading is deterministic and offline: 5,475 decisions, 350 recall-probe tasks, and 81 external-replication calls parse successfully. A prespecified 150-instance rerun used fresh agents and complete per-call logs: it found no retrieval or browser calls and reproduced the named, blinded, and no-evidence pattern (0.760, 0.560, and 0.807). This is a stronger execution check, not a sealed environment: the primary filesystem was not condition-sandboxed, serving and decoding settings were uncontrolled, and network use was prohibited by instruction in the original campaign. Named, blinded, and functional variants change only target and indication fields, keeping the evidence block is byte-identical.

The rerun validates execution only for the logged subset. It does not establish the serving model's weights, rule out all information channels in the primary run, or turn instruction-based network restrictions into a sandbox. We report these constraints because a benchmark audit should make its own execution assumptions visible. Two conditions stopped early and remain exploratory and all other reported clinical conditions cover the full 286 instances.

\paragraph{Interpretation and reporting.} We report accuracy because it is the benchmark's score, but do not interpret it as calibrated probability or scientific utility. The clinical label is imbalanced, so a blinded score near 0.521 must be read against both the majority baseline and the paired named--blinded contrast. For the published tasks, chance values depend on their answer spaces. The key question is whether the above-chance margin persists after identity is removed. We designate the LitQA2 full-passage contrast and its redaction gradient as primary because they directly hold the passage and entity manipulation apart. Other contrasts are included to map failure modes, not to support a single pooled conclusion.

\section{Results}
\subsection{Entity blinding on the clinical-outcome benchmark}
\textsc{blind} replaces the gene, indication, trait, and source names with opaque per-instance tokens, keeping all numeric dossier fields unchanged. With the full dossier, accuracy falls from 0.783 as written to 0.531 [0.474, 0.589], near the 0.521 majority baseline. With identity retained but the dossier removed, it reaches 0.836 [0.788, 0.874]. Thus, in this clinical task, the score is not recovered by the blinded dossier (Figure~\ref{fig:teaser}).

This is a 25.2-point named--blinded gap on the same 286 decisions, not a comparison across separate samples. The result says that the particular rendered dossier does not preserve the performance of the named prompt under opaque-token blinding. It does not say that every numeric field is ignored: the blind transformation removes useful referential and semantic structure together with the entity name. The no-dossier condition also should not be read as a clean ``name-only'' baseline, because its shorter format changes the task. It establishes, however, that high accuracy remains possible without the supplied dossier.

Blinding also removes semantic context, so this result does not show that names simply substitute for evidence. The functional-description condition is therefore diagnostic rather than a clean identity control. That caution is warranted: an adversarial probe recovers the exact HGNC symbol in 114 of 120 descriptions, but in only 2 of 72 blinded prompts:
\begin{center}\small\begin{tabular}{@{}lrrr@{}}\toprule
variant & probes & exact gene recovered & mean confidence \\ \midrule
opaque tokens & 72 & 2/72 & 0.044 \\
functional description & 120 & 114/120 & $\approx0.70$ \\
\bottomrule\end{tabular}\end{center}

\subsection{Recall probe}
The identity result has two explanations: identity may supply biological knowledge or access to the recorded outcome. A no-dossier recall probe asks for the record rather than a prediction and permits \textsc{unknown}. It collects recognition, recorded phase, and programme count for 286 benchmark pairs, 32 real non-clinical pairs, and 32 absent pairs. The outcome and task instruction are removed and opaque identifiers prevent an answer key in task files.

\paragraph{Recall reproduces prediction.} The answered-only comparison is least score-dependent; coding \textsc{unknown} as ``not advanced'' is a sensitivity analysis:
\begin{center}\small\begin{tabular}{@{}lcc@{}}\toprule
 & accuracy & $n$ \\ \midrule
recall, substantive answers only & 0.836 [0.784, 0.878] & 238 \\
no-dossier prediction, same pairs & 0.849 & 238 \\
paired difference & $-0.013$ ($p=0.61$) & \\
answer agreement & 0.937 & \\
recall, all pairs (\textsc{unknown} $=$ not advanced) & 0.818 [0.769, 0.859] & 286 \\
\bottomrule\end{tabular}\end{center}
The recall task produces almost the same answers without supplied evidence. On 218 numeric answers, recalled programme count tracks the database count (Spearman $\rho=0.654$, permutation $p<0.0005$), including within prominence strata. This is a commercial bookkeeping quantity, not something biological plausibility determines. The probe is consistent with access to recorded-outcome information, though it cannot establish what drove any one prediction.

\begin{figure}[t]\centering
\includegraphics[width=0.47\textwidth]{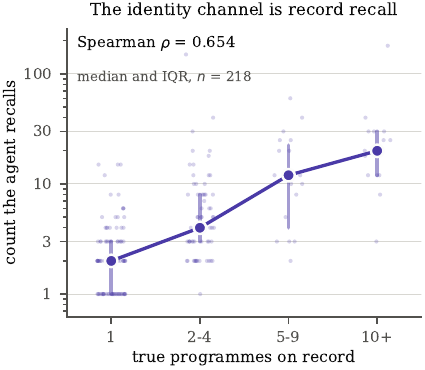}
\caption{\textbf{Recall tracks a database-specific count.} Among 218 numeric recall responses, the recalled number of programmes rises with the number recorded in the source table ($\rho=0.654$). Medians and interquartile ranges are shown.}
\label{fig:recall}\end{figure}

Recognition is reported for 73.4\% of benchmark pairs, 37.5\% of real non-clinical pairs, and 6.3\% of absent pairs, with parallel differences in recalled clinical status. The prespecified instance-level association between recognition and no-dossier prediction is null (Fisher $p=0.37$). A temporal or unpublished-outcome holdout would be preferable, but this dataset cannot support one because every outcome comes from one undated 2023 snapshot. Highest phase is partly public through trial registries, making the programme count the more specific discriminator.

\subsection{Why evidence deletion is not interpretable}
A synthetic line that agrees with the label 95\% of the time yields 0.962 accuracy, while the same line with an independent value yields 0.479. The agent follows the line on 99.0\% and 100\% of tasks, respectively, at similar stated confidence. The positive control therefore detects a directive, not necessarily distributed dossier content. Deleting the dossier raises accuracy from 0.783 to 0.836 (difference $-0.052$, 95\% CI $[-0.085,-0.020]$, $p=0.0026$), but it also changes prompt length, format, and the evidence slot. A swapped-dossier placebo and presentation controls do not identify which component drives the effect, so the placebo substitutes a low-signal genetics block. 

\begin{figure}[t]\centering
\includegraphics[width=0.92\textwidth]{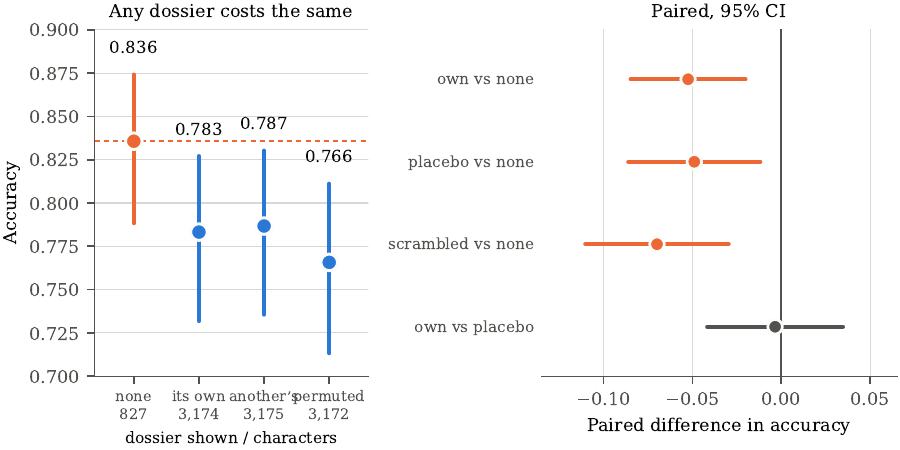}
\caption{\textbf{Deletion does not isolate dossier content.} The own dossier, a matched placebo dossier, and a numerically scrambled dossier all reduce accuracy relative to no dossier. Because the substituted blocks have little independently measured signal, these controls do not identify a content-use effect.}
\label{fig:placebo}\end{figure}

\subsection{Confidence and published benchmark audits}
Across ten conditions spanning 63 points of accuracy, mean stated confidence moves only 12 points (slope +0.068). The agent is under-confident when it performs well and over-confident when blinded:
\begin{center}\small\begin{tabular}{@{}lrrr@{}}\toprule
condition & accuracy & confidence & confidence minus accuracy \\ \midrule
name only & 0.836 & 0.722 & $-0.114$ \\
full dossier & 0.783 & 0.725 & $-0.058$ \\
blinded & 0.531 & 0.639 & $+0.107$ \\
false identity & 0.202 & 0.711 & $+0.509$ \\
\bottomrule\end{tabular}\end{center}

\begin{figure}[t]\centering
\includegraphics[width=0.62\textwidth]{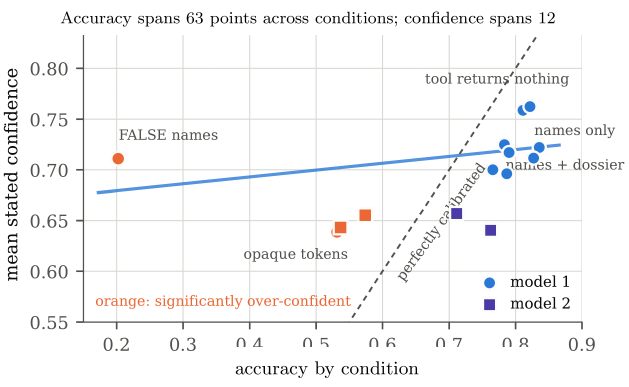}
\caption{\textbf{Stated confidence is weakly related to accuracy across conditions.} Accuracy spans 63 points while mean confidence spans 12. The false-identity condition is particularly over-confident.}
\label{fig:confidence}\end{figure}

Because the own dossier has limited signal, we also audit published tasks that score well above chance. These audits are case studies, not a representative benchmark survey. The four rows below are Biomni Eval1 task families \cite{biomni}; two are vendor LAB-Bench MCQs.\begin{center}\small\begin{tabular}{@{}lccc@{}}\toprule
task family (instances) & prompt supplies & named & blinded \\ \midrule
lab\_bench\_seqqa (21) & the sequences themselves & 0.952 & 1.000 \\
lab\_bench\_dbqa (35) & a database lookup key & 0.629 & 0.171 \\
screen\_gene\_retrieval (50) & written research context & 0.660 & 0.120 \\
gwas\_causal\_gene (147) & name and candidate list & 0.742 & 0.082 \\
\bottomrule\end{tabular}\end{center}
A name-and-candidate-list prompt loses its entire above-chance margin (+66.0 points, $p=3.6\times10^{-26}$; chance 0.093), whereas a sequence task does not move detectably.

The contrast between these rows is useful precisely because their prompt roles differ. In \texttt{lab\_bench\_seqqa}, the sequence remains present after entity blinding and can still determine the answer. In \texttt{lab\_bench\_dbqa}, hiding the lookup key makes the requested record much harder to retrieve. The \texttt{screen\_gene\_retrieval} and \texttt{gwas\_causal\_gene} rows retain the question and candidate set but lose the entity that connects them to a known result. These examples do not identify a universal benchmark defect. Instead, they show why an identity-blinded score is best reported beside the standard score: a large gap is interpretable as a dependency exposed by the intervention, while a small gap is meaningful only when a positive control establishes that the audit could have detected a change.

The sequence result alone is not an equivalence claim: it has one discordant pair. The stronger evidence for a small gap is the 0\% LitQA2 condition below, with 96 questions, no discordant pairs, and $|\mathrm{cost}|\le0.031$. Byte-identical prompts that name no entity give 0.818 versus 0.818, establishing a run-to-run floor. The poor blinded candidate-list performance is consistent with position-only policies, rather than a partial recovery of the named task.

\paragraph{LitQA2 evidence gradient.} LAB-Bench LitQA2 \cite{labbench} provides a key-passage field. We redact the passage in four steps, drawing positions on a token sequence shared across conditions. Entity mentions are never redacted, so the evidence manipulation preserves identity and the blinding manipulation preserves evidence.\begin{center}\small\begin{tabular}{@{}rcccl@{}}\toprule
words redacted & named & blinded & cost & \\ \midrule
0\% & 0.990 & 0.990 & $+0.000$ & no discordant pairs, $|\mathrm{cost}|\le0.031$ \\
33\% & 0.958 & 0.948 & $+0.010$ & $[-0.035,+0.056]$ \\
66\% & 0.865 & 0.771 & $+0.094$ & $[+0.029,+0.159]$, $p=0.012$ \\
100\% & 0.531 & 0.396 & $+0.135$ & $[+0.029,+0.242]$, $p=0.024$ \\
\bottomrule\end{tabular}\end{center}
Within this fixed masking scheme, the identity gap grows as evidence is removed (per-instance slope +0.147 per unit redaction, $p=0.004$). The 66\% level matters because neither condition is near ceiling (0.865 versus 0.771): the vanishing gap at 0\% is consistent with sufficient evidence, not merely saturation. The 66-point collapse reproduces on the second model at +62.6 points. Up to ten instances share an agent call, so we report call-clustered resampling beside paired McNemar inference \cite{cgm}.

\subsection{Robustness and interpretation}
For the named-versus-no-evidence contrast, the paired i.i.d. standard error is 0.0165 and the paired cluster-robust standard error is 0.0133; an unpaired clustered analysis gives 0.0309 because it discards within-instance pairing. We retain paired inference as primary, but do not make a global multiple-testing claim over roughly 40 tests.

The narrowest claim supported by the primary analysis is an interaction within one fixed LitQA2 masking scheme: as more words in the supplied passage are redacted, the named--blinded gap increases. We need to note however, that it does not estimate the prevalence of record recall among agents or among life-science benchmarks, nor does it prove that every named response used identity. Those stronger causal questions would need new benchmark construction, like temporally held-out outcomes, independently verified aliases, or a sealed inference environment. We include the clinical and external case studies to make the design limitations concrete, rather than to substitute them for such experiments.

Four clinical conditions were repeated over all 286 instances on the smaller model with byte-identical prompts. Dispatch-mode checks on twelve blinded batches gave the same functional-versus-blinded gap (+0.071) in one-call and grouped execution. These are useful sensitivity checks, but not independent replication across model families. Similarly, functional descriptions are not a clean blinding control, since they omit identifiers by instruction but can still reveal a gene through a distinctive biological description.

\section{Discussion and conclusion}
The audit measures what a score depends on under a specified intervention, not whether an agent can reason in general. In our clinical-outcome benchmark, the blinded dossier does not recover the named score, and the recall probe is consistent with access to recorded outcomes. In external case studies, the identity gap varies with how much decisive evidence the prompt supplies. These results motivate a discriminant-validity check for benchmarks intended to operationalise reasoning over evidence \cite{jacobs}, as they do not estimate how common identity dependence is across the field. A bare evidence-deletion result cannot establish content use.

\paragraph{Recommendations.} Benchmark builders should consider reporting an identity-blinded comparison, retrieval telemetry where tools are available, and confidence by condition. A positive control can make a null more interpretable, a placebo should replace signal rather than noise, and public outcome tables merit a direct recall probe.
\subsection{Limitations}
The identity result cannot separate recognition from the possibility that opaque tokens hurt the agent's reasoning, so +25.9 would be an upper bound. Whether prediction used recall on a given instance has a null prespecified test (Fisher $p=0.37$) and an exploratory rebuild that we do not treat as a claim. The deletion effect is bounded, not explained. Our claims cover audited entity-level tasks with snapshot-level, censored, undated labels. On execution: two models from one family, one harness, neither certifiable, and decoding uncontrolled.

\end{document}